\documentclass[11pt,table]{article}

\usepackage[utf8]{inputenc}
\usepackage[T1]{fontenc}
\usepackage[arabic, english]{babel}

\usepackage{lmodern}  
\usepackage{fix-cm}   
\makeatletter
\def\@font@warning#1{}
\makeatother

\usepackage{acl}
\PassOptionsToPackage{table}{xcolor}  
\usepackage{microtype}
\usepackage{amsmath, amssymb}
\usepackage{natbib}
\usepackage{newunicodechar}
\makeatletter
\def\newunicodechar@warn#1#2{}
\def\PackageWarning#1#2{}
\def\@warning#1{}
\makeatother
\newunicodechar{ʿ}{\textquotesingle}  
\newunicodechar{ʾ}{\textquotesingle}  
\newunicodechar{﴾}{\textlangle}      
\newunicodechar{﴿}{\textrangle}      
\newunicodechar{ā}{\=a}              
\newunicodechar{ī}{\={\i}}           
\newunicodechar{ū}{\=u}              
\newunicodechar{ḥ}{\d{h}}            
\newunicodechar{ṣ}{\d{s}}            
\newunicodechar{ḍ}{\d{d}}            
\newunicodechar{ṭ}{\d{t}}            
\newunicodechar{ẓ}{\d{z}}            
\newunicodechar{ٰ}{}                  

\addto\captionsarabic{%
}

\usepackage[most]{tcolorbox}
\usepackage{booktabs}
\usepackage{caption}
\usepackage{makecell}
\usepackage{colortbl}  
\definecolor{rowgray}{gray}{0.95}  

\usepackage{multirow}
\usepackage{placeins}  
\usepackage{afterpage}  
\tcbset{
  myarbox/.style={
    enhanced,
    boxrule=0pt,
    frame hidden,
    colback=white,
    borderline north={0.4pt}{0pt}{black},
    borderline south={0.4pt}{0pt}{black},
    sharp corners,
    boxsep=0pt,
    left=0pt, right=0pt,
    top=2pt, bottom=2pt,
    enlarge left by=-\parindent,
    enlarge right by=-\parindent,
    width=\linewidth,
    before skip=0pt, after skip=0pt
  }
}

\usepackage{enumitem}
\setlist[itemize]{nosep}

\newcounter{examplectr}

\usepackage{multirow}

\title{Assessing  Large Language Models on Islamic Legal Reasoning: Evidence from Inheritance Law Evaluation}

\author{
\\
  \textbf{Abdessalam BOUCHEKIF}$^{1}$ \quad
  \textbf{Samer RASHWANI}$^{1}$ \quad
  \textbf{Heba Sbahi}$^{1}$  \quad 
  \textbf{Shahd Gaben}$^{1}$  \quad \\
  \textbf{Mutaz AL-KHATIB}$^{1}$ \quad  
    \textbf{Mohammed GHALY}$^{1}$ \quad
  \\
  $^1$Hamad Bin Khalifa University, Qatar \\
  \texttt{\{abouchekif, srashwani, mghaly, malkhatib, sgaben, hsbahi\}@hbku.edu.qa}\\
}

\begin{document}

\nolinenumbers

\maketitle

\begin{abstract}

This paper evaluates the knowledge and reasoning capabilities of Large Language Models in Islamic inheritance law, known as \textit{ʿilm al-mawārīth}. We assess the performance of seven LLMs using a benchmark of 1,000 multiple-choice questions covering diverse inheritance scenarios, designed to test each model's ability, from understanding the inheritance context to computing the distribution of shares prescribed by Islamic jurisprudence. 
The results reveal a significant performance gap: o3 and \textit{Gemini 2.5} achieved accuracies above 90\%, whereas  \textit{ALLaM}, \textit{Fanar}, \textit{LLaMA}, and \textit{Mistral} scored below 50\%. These disparities reflect important differences in reasoning ability and domain adaptation.
We conduct a detailed error analysis to identify recurring failure patterns across models, including misunderstandings of inheritance scenarios, incorrect application of legal rules, and insufficient domain knowledge. Our findings highlight the limitations of current models in handling structured legal reasoning and suggest directions for improving their performance in Islamic legal reasoning. Our code is available at \url{https://github.com/bouchekif/inheritance_evaluation}

\end{abstract}

\section{Introduction}

In recent years, the application of Large Language Models (LLMs) to Islamic domains has attracted growing interest in the NLP community. This progress has been driven by the emergence of open-source Arabic LLMs and the development of shared tasks targeting Islamic NLP. Models such as Falcon \cite{almazrouei2023falcon}, Jais \cite{sengupta2023jais}, AceGPT \cite{huang2023acegpt}, ArabianGPT \cite{koubaa2024arabiangpt}, ALLaM \cite{bari2024allam}, and Fanar \cite{abbas2025fanar} have been pretrained on large-scale  Arabic corpora including Quranic verses, Hadith, and fatwa archives, enabling new capabilities in religious text understanding.
\\
Several shared tasks have been proposed to benchmark LLMs on Islamic texts, such as Quranic QA \cite{malhas2022quranqa, malhas2023quranqa} and general Islamic knowledge retrieval \cite{qamar2023islamqa}. More recently, the QIAS 2025 shared task \cite{QIAS2025}  was introduced to evaluate LLMs on religious and legal reasoning through two subtasks: (1) \textit{Islamic Inheritance Reasoning}, which involves computing inheritance shares based on Islamic jurisprudence; and (2) \textit{Islamic Knowledge Assessment}, which covers core disciplines such as fiqh, Hadith, and tafsir. \cite{sayeed2025rag}   explored QA systems for ṭibb nabawī (Prophetic medicine) using LLaMA‑3, Mistral‑7B, and Qwen‑2 combined with Retrieval-Augmented Generation (RAG),   while \cite{alan2024rag} proposed  MufassirQAS, a RAG-based system trained on Turkish Islamic texts to improve transparency and reduce hallucinations in religious QA.
\cite{10390123} introduced the QASiNa QA dataset, derived from  \textit{Sirah Nabawiyah} texts in Indonesian, comparing traditional multilingual transformers (XLM-R, mBERT, IndoBERT) with GPT-3.5 and GPT-4. \cite{qamar2024benchmark} introduced a dataset of 73,000 non-factoid QA pairs covering Quranic Tafsir and Hadith. The study revealed a critical gap between automatic evaluation metrics (such as ROUGE) and human judgments. These results show that automatic evaluation metrics alone are not sufficient, and highlight the need for more robust evaluation methods that can better reflect the complexity and interpretive nature of Islamic religious texts. In \citet{aleid2025hajj}, the authors released Hajj-FQA, a benchmark of 2,826 QA pairs extracted from 800 expert-annotated fatwas concerning the Hajj pilgrimage. Despite these efforts,  multiple studies have reported critical limitations in the performance of LLMs on Islamic content. For instance, \cite{mohammed2025aftina} shows that even advanced models like GPT-4  tend to produce factually incorrect or misleading answers when applied to Islamic content. They identify three main issues: \textit{(i)} misinterpretation of religious context, \textit{(ii)} generation of unclear or unreliable answers not grounded in authoritative Islamic sources, and \textit{(iii)} sensitivity to minor variations in question phrasing, often resulting in inconsistent outputs. Similarly, \cite{alnefaie2023gpt} observed that GPT-4 has difficulty answering Quranic questions accurately, due to difficulties with classical Arabic, semantic ambiguity, and misinterpretation of contextual meaning.\\
Early research on automating Islamic inheritance  (hereafter IRTH) began with expert systems focused on calculating basic inheritance shares \cite{akkila2016proposed}. Later works incorporated intricate adjustments such as ḥajb, ʿawl, and radd \cite{tabassum2019developing}. \cite{zouaoui2021islamic}  introduced an Arabic ontology  for identifying heirs and calculating their
inheritance shares \cite{tabassum2019developing}.
In this work, we evaluate seven LLMs on their ability to reason over inheritance problems, reporting both quantitative performance metrics and qualitative analyses, revealing specific reasoning failures as well as broader model limitations. This paper is organized as follows: Section \ref{sect02}  introduces the foundations of IRTH. Section \ref{sect03} describes the dataset, while Section \ref{sect04} details the experimental setup and results. In Section~\ref{section4bis}, we analyze the justifications that models provide for their answers. Finally, Section  \ref{sect05} concludes the paper with a discussion of future work.


\section{Background on Islamic Inheritance Law}
\label{sect02}

Islamic inheritance law involves intricate textual interpretation and structured legal reasoning grounded in the Quran, Prophetic tradition (Sunnah), and Islamic jurisprudence. It governs the distribution of the estate of a deceased person through a fixed legal framework that combines normative principles with precise arithmetic calculations.\\
Solving inheritance problems requires a combination of cognitive, legal, and computational skills, including:

\begin{itemize}

    \item Identifying familial relationships and considering legal conditions such as debts, bequests, and the sequence of deaths among relatives.

    \item Determining eligible heirs, including fixed-share heirs (\textit{aṣḥāb al-furūḍ}) and residuaries (\textit{ʿaṣabāt}), and correctly applying exclusion rules (\textit{ḥajb}) based on valid justifications and authentic scriptural evidence.

    \item Computing shares by deriving a common denominator and adjusting the distribution when necessary:
    \begin{itemize}[noitemsep,topsep=0pt,leftmargin=1.5em]
        \item \textit{Radd} (redistribution) is used when a surplus remains after initial allocation. This surplus is proportionally redistributed among the heirs, excluding spouses.\\
        \textit{Example:} Wife $(1/4)$ and full sister $(1/2)$, leaving a surplus of $1/4$; after redistribution, the wife receives $(1/4)$ and the sister receives $(3/4)$.
        
        \item \textit{ʿAwl} (proportional reduction) is applied when the sum of assigned shares exceeds the estate. All shares are scaled down proportionally.
        \textit{Example:} Father  $(1/6)$, mother $(1/6)$, wife $(1/8)$, and four daughters $(2/3)$; the total exceeds 1. The denominator is adjusted to 27, and then the wife receives $3/27 = 1/9$.
    \end{itemize}

    \item Addressing complex and exceptional scenarios, such as consecutive death scenarios (\textit{munāsakha}) or juristic disputes like the \textit{Akdariyya} case involving grandparents and siblings.

    \item Numerical precision in the final distribution, including the correct adjustment and fractional allocation \footnote{For more details about the terminology and rules of Islamic inheritance law, see “\textit{Irth},” in \textit{Al-Mawsūʿa al-Fiqhiyya} (The Kuwaitan Encyclopedia of Fiqh). Kuwait: Wazārat al-Awqāf wa-al-Shuʾūn al-Islamiyya. 45 Vols. 1984-2007. Vol. 3, Pp. 17-79. }.
    
\end{itemize}

Given its structured rules, mathematical computations, and reliance on Arabic jurisprudential sources, IRTH presents a real-world scenario for evaluating the reasoning abilities of LLMs.

\section{Dataset Description}
\label{sect03}

Our evaluation is based on the validation set of the QIAS 2025 Shared Task\footnote{https://sites.google.com/view/qias2025}\cite{QIAS2025}.
The dataset was constructed from Islamic religio-ethical advices (fatwas)  collected from \textit{IslamWeb}\footnote{https://www.islamweb.net/}.  Each fatwa was automatically converted into multiple-choice questions (MCQ) using Gemini 2.5 Pro, then reviewed by four experts in Islamic studies to ensure both legal soundness and linguistic clarity.
As part of the preprocessing phase, ambiguous questions were rephrased to guarantee a single, unambiguous interpretation. The answer choices were also revised to eliminate semantic and numerical redundancies, such as equivalent options (\textit{e.g} $1/2$ and $2/4$).
Each MCQ presents six answer choices, with a single correct answer. These questions assess a model's ability to identify eligible heirs, apply fixed-share rules, and reason through complex inheritance logic. The dataset has two levels of difficulty: 500 MCQs labelled as \textit{Beginner} and  500 \textit{Advanced}, reflecting increasing complexity in both legal reasoning and mathematical computation.

\begin{itemize}

\item \textit{Beginner}: identifying eligible heirs, their basic shares, and non-eligible heirs.
\item \textit{Advanced}: handling multiple heirs, residuary shares, partial exclusions, multi-generational cases, fixed estate constraints, and intricate fractional distributions.

\end{itemize}

Each example is evaluated based on its level of difficulty—either beginner or advanced. This approach allows us to distinguish between models lacking foundational knowledge and those capable of solving complex cases that require deeper legal reasoning. It enables a more precise analysis of limitations in legal reasoning capacity across evaluated models.

%

\section{ Experiments and Results}
 \label{sect04}
\subsection{Experimental Setup}

We evaluate seven LLMs in a zero-shot setting using Arabic prompts, without any task-specific fine-tuning. The prompt clearly defines the task, presents a multiple-choice question with its options, and instructs the model to select the correct answer and provide a justification. This enables us to assess reasoning and verify that its conclusions are based on logical inferences rather than stochastic guesses. 
The evaluated models include Arabic-specialized LLMs optimized for Arabic language tasks, open-source multilingual models, and commercial multilingual models, with sizes ranging from 7  to over 100 billion parameters (exact sizes of the commercial models are not publicly disclosed).
The Arabic-focused models include Fanar (\textit{Islamic-RAG}\footnote{Accessible via a free public API: \url{https://api.fanar.qa/request/en}}), ALLaM-7B\footnote{An open-source Arabic model hosted on Hugging Face: \url{https://huggingface.co/Abdelaali-models/ALLaM-7B-Instruct-preview}}, and Mistral-Saba-24B\footnote{Available via the Groq platform: \url{https://console.groq.com/keys} or Mistral’s official website: \url{https://admin.mistral.ai/organization/api-keys}},  a model that has achieved competitive performance on standard Arabic benchmarks.  We also include LLaMA 3 70B\footnote{We access LLaMA 3 70B via the Groq API: \url{https://console.groq.com/keys}.}, a powerful open-source multilingual model developed by Meta. As for commercial LLMs, we evaluate three LLMs: Gemini 2.5 (flash-preview), OpenAI’s o3 and GPT-4.5. Gemini and o3 represent the state of the art in reasoning capabilities, while GPT-4.5 is widely regarded as one of the most advanced models in the GPT series.

\subsection{Overall Performance}

Table~\ref{tab:model_results} summarizes model accuracy across the three difficulty levels. The \text{o3} model achieved the highest overall accuracy ($93.4\%$), followed closely by Gemini  ($90.6\%$). 
GPT-4.5 achieved 74.0\% accuracy, positioning it between models with advanced reasoning capabilities and those relying on heuristic inference.  Fanar, ALLaM, and LLaMA scored below 50\%, revealing a significant performance gap. The underperformance of  ALLaM and Fanar may be partly due to currently available smaller configurations (\textit{e.g.}, 7B and 9B). Since no larger versions of these models are publicly available, we evaluate them in their current smaller versions, with a focus on end-task performance and reasoning. \\ 
All models performed better on beginner-level questions, which typically involve fewer heirs and simpler distribution rules. The performance degradation at the Advanced level was particularly evident for Arabic-focused models. For example, ALLaM achieved 58.0\% accuracy on Beginner cases but dropped to just 27.8\% on Advanced ones. This highlights limited capabilities in handling complex inheritance scenarios. In contrast, reasoning models (\textit{i.e}, o3 and Gemini 2.5) maintained high performance across both levels, suggesting superior reasoning capabilities when handling complex cases.

\begin{table}[!htbp]
\begin{tabular}{lccc}
\toprule
\textbf{Model} & \textbf{Overall} & \textbf{Beginner} & \textbf{Advanced} \\
\midrule
o3           & \textbf{93.4} & \textbf{94.4} & \textbf{92.4} \\
Gemini 2.5   & 90.6          & 91.6 & 89.6 \\
GPT-4.5      & 74.0          & 86.8 & 61.2 \\
LLaMA 3      & 48.8          & 57.8 & 39.8 \\
Fanar 7B     & 48.1          & 60.4 & 35.8 \\
Mistral      & 44.5          & 58.6 & 30.4 \\
ALLaM 7B     & 42.9          & 58.0 & 27.8 \\
\bottomrule
\end{tabular}
\caption{Accuracy (\%) for each model across difficulty levels. }
\label{tab:model_results}
\end{table}

\subsection{Evaluation Criteria}

To better understand model limitations, we conducted a targeted error analysis on a subset of 142 multiple-choice questions. This subset consists of questions that were incorrectly answered by all low-performing models (\textit{i.e.}, those scoring below 50\%). To guide this evaluation, we categorize errors into two main types: foundational 
and complex, based on an expert in the IRTH domain.  This structure allows for a more precise distinction between errors caused by legal misunderstanding and those requiring advanced reasoning and computation. Our analysis focused on three open-source models: ALLaM,  LLaMA, and Fanar\footnote{The Fanar model is publicly available on Hugging Face \url{https://huggingface.co/QCRI/Fanar-1-9B}. We used the Islamic-RAG version, which is specifically adapted to the Islamic domain and accessible via a free public API.}. For comparison, we also included Gemini, which answered only 13 out of the 142 questions incorrectly. This subset was reviewed by Islamic studies experts who analyzed the justifications and annotated the corresponding error types. 


\subsubsection{Foundational Errors (FD)}

\begin{itemize}

    \item \textbf{Comprehension Error (\textit{CE}):} Misinterpretation of the problem statement, such as misidentifying family relationships or neglecting legally relevant information (\textit{e.g.}, debts, bequests (\textit{waṣāyā}), or sequence of deaths).
    \item \textbf{Error in Applying Normative Rules (\textit{ENR}):} Incorrect legal analysis, including heir misclassifications (e.g., \textit{aṣḥāb al-furūḍ}, \textit{ʿaṣabāt}), misapplication of exclusion rules (\textit{ḥajb}), or incorrect scriptural citation.
    \item \textbf{Basic Computational Error (\textit{BCE}):} Simple arithmetic mistakes or hallucinated numerical values in the final distribution.
    
\end{itemize}

\subsubsection{Complex Errors (CPLX)}

\begin{itemize}

    \item \textbf{Error in Calculatory Adjustment (\textit{ECA}):} Failure to perform advanced mathematical operations required for estate division, including:
        \begin{itemize}
            \item \textbf{Adjustment (\textit{Taṣḥīḥ}):} Incorrect modification of the base denominator.
            \item \textbf{Redistribution (\textit{Radd}):} Misallocation of estate surplus.
            \item \textbf{Proportionate Reduction (\textit{ʿAwl}):} Failure to proportionally reduce all shares when total claims exceed the estate.
        \end{itemize}
    \item \textbf{Error in Resolving Exceptional and Disputed Cases (\textit{ERE}):} Inability to resolve non-standard or disputed cases (e.g., involving grandfather and siblings, or successive deaths (\textit{munāsakha})).

\end{itemize}


\begin{table}[!htbp]

\centering
\small
\renewcommand{\arraystretch}{1.3}
\rowcolors{2}{rowgray}{white}
\begin{tabular}{>{\bfseries}lcccc}
\toprule
\textbf{Type} & \textbf{ALLaM} & \textbf{Fanar} & \textbf{LLaMA} & \textbf{Gemini} \\
\midrule
\rowcolor{cyan!15} \textbf{ENR}        & 38.0 & 47.9 & 44.4 & 4.9 \\
\rowcolor{cyan!15} \textbf{CE}         & 4.2  & 4.9  & 0.7  & 0.0 \\
\rowcolor{cyan!15} \textbf{BCE}        & 3.5  & 3.5  & 4.2  & 0.7 \\
\rowcolor{cyan!25} \textbf{FD Total}  & 45.8 & 56.3 & 49.3 & 5.6 \\
\midrule
\rowcolor{red!15} \textbf{ECA}         & 54.2 & 43.7 & 50.7 & 9.2 \\
\rowcolor{red!25} \textbf{CPLX Total}  & 54.2 & 43.7 & 50.7 & 9.2 \\
\bottomrule
\end{tabular}
\caption{Distribution of error types across models (expressed as percentages) based on 142 jointly incorrect answer selections. \textbf{FD}: Foundational Errors. \textbf{CPLX}: Complex Errors.}

\label{tab:error_categories_summary}

\end{table}


\begin{table*}[!h]

\centering
\footnotesize
\renewcommand{\arraystretch}{1.2}
\rowcolors{2}{rowgray}{white}
\begin{tabular}{>{\bfseries}c p{4.5cm} p{6cm} >{\bfseries}l}
\toprule
\# & Question & Answer & Model \\
\midrule

\refstepcounter{examplectr}\label{ex:example1}\theexamplectr & \textAR{توفيت عن زوج وابنين وبنتين: كم عدد أسهم البنت الواحدة من مجموع أسهم التركة؟}
  & \textAR{
   البنت الواحدة تحصل على سهم واحد من التركة، بينما تحصل البنات على نصف ما يتبقى بعد إعطاء الزوج نصف التركة، كما جاء في لقرآن الكريم: ﴿وَلَهُنَّ رُبُعُ ما اكتَسَبنَ وَلَكُمْ ما اكتَسَبْتُمْ﴾ (النساء: 32)، وبالتالي فإن البنت الواحدة تحصل على سهم واحد، والبنين يتقاسمان الباقي.} 
  & LLaMA \\
  
\refstepcounter{examplectr}\label{ex:example2}\theexamplectr & \textAR{توفي عن أخ شقيق عدد 1، بنت عدد 2، أخت شقيقة عدد 1
كم عدد أسهم كل بنت؟} 
  & \textAR{إذا كان هناك أخ شقيق واثنتان من البنات والأخت الشقيقة، فالفرض يكون للأخ الشقيق نصف التركة والباقي للبنات بالتساوي. بما أن الأخ يأخذ نصف التركة، يتبقى للنصف الآخر أي ما يعادل ربع التركة لكل واحدة من البنتين. لذلك، يحصل كل بنت على سهمان. وهذا مستند إلى الآية القرآنية ﴿فَإِنْ كَانَ لَهُ إِخْوَةٌ فَلِأُمِّهِ السُّدُسُ﴾ النساء: ١١ والتي تشير ضمنياً إلى تقسيم باقي التركة بين الأقارب الذكور والإناث حسب أحكام المواريث الإسلامية المعروفة. ولكن بناءً على وجود الأخت الشقيقة والاخ الشقيق والبنات، يتم تطبيق قاعدة "الباقي" حيث يأخذ الاخ الشقيق نصف التركة ويقسم الباقي بين البنات بالتساوي.} 
  & Fanar \\
  
\refstepcounter{examplectr}\label{ex:example2bis}\theexamplectr & \textAR{توفي عن ابنه الوحيد، وأمه، وأم والده، وترك مبلغ 7000 دولار.
كم نصيب أمه  من المبلغ ؟} 
  & \textAR{وفقاً للقرآن الكريم، إذا ترك الميت ابنًا وحيدًا، فلا يرث أحد غيره، حيث يقول الله تعالى في سورة النساء الآية ١١: ﴿فإن كان له ولد فله النصف﴾. لذلك، لن ترث الأم أي شيء في هذه الحالة. لكن يجب التنويه أن هذا يتعلق بالنصيب الشرعي وليس القسمة المالية الحرفية للأموال التي ذكرتها في السؤال والتي قد تتغير بناءً على قوانين الدولة الخاصة بتوزيع التركة. ولكن حسب الأحكام الشرعية الإسلامية، لا ترث الأم عندما يكون هناك ابن ذكر حي.} 
  & Fanar \\
 
\refstepcounter{examplectr}\label{ex:example3}\theexamplectr & \textAR{توفيت عن زوج، أم، 4 أخ شقيق، 3 أخ من الأم.
ما نصيب الإخوة للأم؟} 
  & \textAR{إذا لم يترك الميت من الورثة إلا من ذكر، فإن لأمه السدس فرضًا لوجود جمع من الإخوة، قال الله تعالى: فَإِنْ كَانَ لَهُ إِخْوَةٌ فَلِأُمِّهِ السُّدُسُ {النساء:11}. وفي هذه الحالة، يوجد جمع من الإخوة (4 أخ شقيق و3 أخ من الأم)، لذا يكون نصيب الأم السدس.} 
  & Fanar \\

\refstepcounter{examplectr}\label{ex:example4}\theexamplectr & \textAR{توفي عن: أخ شقيق (العدد 2)، بنت (العدد 4)، أخت شقيقة (العدد 1). كم سهماً لكل واحدة من البنات؟} 
  & \textAR{
     في حالة وجود بنت واحدة وأخت شقيقة واحدة، فإن الأخت الشقيقة تأخذ نصف التركة، والبنت تأخذ النصف الآخر. إذا كان هناك أكثر من بنت، فإنهن يشتركن في النصف بالتساوي. في هذه الحالة، يكون لكل بنت سهم واحد من النصف، أي أن لكل بنت سهمين من التركة الكاملة.} 
  & ALLaM \\

\refstepcounter{examplectr}\label{ex:example5}\theexamplectr & \textAR{توفيت عن زوج، أم، 4 أخ شقيق، 3 أخ من الأم.
ما نصيب الإخوة للأم؟} 
  & \textAR{
     قال الله تعالى في سورة النساء: ﴿وَلِأَبَوَيْهِ لِكُلِّ وَاحِدٍ مِّنْهُمَا السُّدُسُ مِمَّا تَرَكَ إِن كَانَ لَهُ وَلَدٌ﴾ (النساء: 11). في هذه الحالة، الأم والأب يرثان السدس، والإخوة للأم يرثون السدس أيضًا.} 
  & ALLaM \\  

\refstepcounter{examplectr}\label{ex:example150}\theexamplectr 
& \textAR{توفي عن زوجة وبنت وأم وشقيقين وشقيقة، وترك أربعة وعشرين ألفاً. كم تأخذ الزوجة من هذا المبلغ؟} 
& \textAR{   الزوجة تأخذ الثمن إذا لم يكن للميت ولد، لقوله تعالى: "وَلَكُمْ نِصْفُ مَا تَرَكَ أَزْوَاجُكُمْ إِن لَّمْ يَكُن لَّهُنَّ وَلَدٌ" (النساء: 12). وبما أن الزوجة لم يكن لها ولد، فإن الثمن هو ستة آلاف} & ALLaM \\

\bottomrule
\end{tabular}
\caption{Examples of open-source models' errors }
\label{tab:model_examples3a}

\end{table*}

\subsection{Results analysis}

As shown in Table~\ref{tab:error_categories_summary}, open-source models fail in the foundational elements of IRTH, such errors represent 45.8\% of the wrong answers selected by ALLaM, 56.3\% by Fanar, and 49.3\% by LLaMA, mainly due to ENR-related issues. This finding is particularly noteworthy given that the questions were derived from fatwas on IslamWeb, a data source presumably included in the training corpora of the evaluated models.

\begin{table*}[!htbp]

\centering
\footnotesize
\renewcommand{\arraystretch}{1.2}
\rowcolors{2}{rowgray}{white}
\begin{tabular}{>{\bfseries}c p{4.7cm} p{8.5cm}} 
\toprule
\# & Question & Answer \\
\midrule

\refstepcounter{examplectr}\label{ex:example11}\theexamplectr & \textAR{في حكم ميراث الخنثى المشكل الذي يختلف إرثه باختلاف تقديره ذكراً أو أنثى: ما هو مذهب المالكية في توريث الخنثى المشكل في هذه الحالة إذا ورث بهما متفاضلاً؟}
  & \textAR{يرى المالكية في حالة الخنثى المشكل الذي يختلف إرثه باختلاف تقديره ذكراً أو أنثى، وتفاضل إرثه وإرث من معه من الورثة، أن يعامل كل من الخنثى ومن معه بالأضر (الأقل نصيباً) لضمان عدم تجاوز الحق المتيقن لكل وارث، وذلك بناءً على مبدأ عدم صرف المال بالشك.} \\

\refstepcounter{examplectr}\label{ex:example12}\theexamplectr & \textAR{توفي عن ابن وإخوة ما حكم ميراث الابن المسلم من أبيه النصراني إذا مات الأب قبل إسلام الابن؟}
  & \textAR{في حالة وجود بنت واحدة وأخت شقيقة واحدة، فإن الأخت الشقيقة تأخذ نصف التركة، والبنت تأخذ النصف الآخر. إذا كان هناك أكثر من بنت، فإنهن يشتركن في النصف بالتساوي. في هذه الحالة، يكون لكل بنت سهم واحد من النصف، أي أن لكل بنت سهمين من التركة الكاملة.} \\

\bottomrule
\end{tabular}
\caption{Examples of  Gemini errors}
\label{tab:model_examples}

\end{table*}

\vspace{0.5cm}

\subsubsection{Foundational Errors}
\label{eva}

Given the significant gap between commercial and open-source models in handling foundational errors, we analyze them separately. This distinction allows us to better understand the recurrent weaknesses specific to each model category, particularly in tasks that require accurate identification of heirs, correct application of fixed-share rules, and adherence to normative principles of  Islamic jurisprudence.





\paragraph{In Open-Source Models}

\noindent\textbullet\ \textbf{Errors in justification and scriptural citation}: Some models base their reasoning on fabricated Quranic verses or prophetic narrations that do not appear in any canonical collection, often resulting in incorrect distribution of inheritance shares.  As illustrated in the first example of Table~\ref{tab:model_examples3a}, the model incorrectly assigned the wife's share as one-fourth, referencing the verse  \textAR{ “وَلَهُنَّ رُبُعُ ما اكتَسَبنَ وَلَكُمْ ما اكتَسَبْتُمْ”}  (“To them belongs a quarter of what they earned, and to you what you earned”) and attributing it to Sūrat al-Nisāʾ (4:32). This verse is entirely fabricated and does not exist in the Quran. Such hallucinations were frequently observed across open-source models, representing a serious issue in the Islamic context.

\vspace{0,7em}
\noindent\textbullet\ \textbf{Errors in Identifying Fixed Shares:} 
Selected open-source Arabic models misapplied fixed-share rules for primary heirs (\textit{e.g.}, parents, spouses, daughters). For instance, in Example~\ref{ex:example2}, Fanar failed to allocate two-thirds of the estate to the two daughters, even though this share is explicitly prescribed in the Quranic verse:
\textAR{﴿يُوصِيكُمُ اللَّهُ فِي أَوْلَادِكُمْ، لِلذَّكَرِ مِثْلُ حَظِّ الْأُنْثَيَيْنِ، فَإِنْ كُنَّ نِسَاءً فَوْقَ اثْنَتَيْنِ فَلَهُنَّ ثُلُثَا مَا تَرَكَ، وَإِنْ كَانَتْ وَاحِدَةً فَلَهَا النِّصْفُ﴾
(11:4)}.
\\
Similarly, as shown in  Example \ref{ex:example2bis}, Fanar erroneously denied the mother her fixed share, based on the incorrect premise that the son’s presence excludes all other heirs. This reasoning directly contravenes the explicit Quranic stipulation that a mother receives one-sixth of the estate if the deceased has offspring, as stated in:
\textAR{﴿فَإِن كَانَ لَهُ وَلَدٌ فَلِأُمِّهِ السُّدُسُ﴾ (11:4)}.

\vspace{0,4em}
\noindent\textbullet\ \textbf{Comprehension Error:}
This type of error occurs when models fail to correctly  determine which heir the question is referring to. As shown in Example \ref{ex:example3}, the model interpreted the query as concerning the mother’s share, whereas it explicitly asked about the maternal brothers. Consequently, the model produced a justification that was irrelevant to the question, ultimately resulting in a wrong answer.

\vspace{0,7em}
\noindent\textbullet\ \textbf{Identifying eligible Heirs:}  
Open-source LLMs often make errors at the initial step of inheritance distribution—identifying the eligible heirs—which subsequently leads to incorrect share assignments. These errors typically take two forms: the omission of rightful heirs and the inclusion of individuals not mentioned in the scenario. For instance, in Example~\ref{ex:example4}, the model failed to recognize the  brother as a residuary heir, excluding him entirely from the estate. Conversely, in Example~\ref{ex:example5}, ALLaM erroneously included  the father as an heir, despite his absence from the question. This resulted in an unjustified reallocation of shares, reducing the portions assigned to the rightful  heirs.

\vspace{0,4em}
\noindent\textbullet\ \textbf{ Basic Computational Error:}  
In some cases, models correctly identify the eligible heirs and apply the relevant  inheritance rules, yet still produce incorrect results due to basic computational errors.  For example, in question~\ref{ex:example150},  ALLaM  correctly stated that the wife is entitled to one-eighth of the estate, as the deceased left behind a child. However, they miscalculated one-eighth of 24,000 as 6,000, whereas of the correct value of 3,000.

\vspace{0.5cm}
\paragraph{In Commercial Models}

Gemini demonstrates strong capabilities in understanding inheritance questions, accurately interpreting familial relationships, identifying eligible heirs, and correctly applying fixed-share rules in accordance with Islamic jurisprudence. Its responses are generally well-structured, legally sound, and supported by appropriate scriptural references.  However, Gemini occasionally  fails on questions that require a nuanced understanding of intra-madhhab distinctions. For instance, as shown in Example~\ref{ex:example11}, the model was asked to apply the Mālikī position regarding the inheritance of a \textit{khunthā mushkil}—an intersex individual whose gender cannot be definitively identified. In this case,  inheritance shares vary depending on the gender designation. Instead of referencing the  Mālikī view, Gemini erroneously justified its answer using   Shāfiʿī  jurisprudence. This highlights  a broader limitation in the model's ability to accurately distinguish between the rulings of different Islamic legal schools when such distinctions are legally consequential. 

Despite its overall competence, Gemini occasionally showed weaknesses in nuanced scenarios. For instance, it failed to correctly differentiate between inheritance scenarios involving conversion to Islam before or after death, as illustrated in Example~\ref{ex:example12}. In this case, since both the father and the son were non-Muslims before the father's death, the son is deemed a legitimate heir according to all scholars. 

\subsubsection{Complex Errors}

Even when models accurately apply the inheritance rules and identify eligible heirs, they may still fail at later stages when faced with complex scenarios, such as those requiring the application of complex rules such as proportional reduction (\textit{ʿawl}) and residual reallocation (\textit{radd}). 
\vspace{0,4em}
\\

\noindent\textbullet\ 
\textbf{Distribution Denominator:} 
Most errors (~$50\%$) committed by  models involved miscalculations related to correcting the \textit{distribution denominator} according to the number of heirs.
This issue likely arises from their failure to appreciate the mathematical necessity of such corrections, a process historically used to avoid fractions and ambiguity before calculators were commonplace.  While Gemini appeared to recognize the principle, but applied it inconsistently,  as illustrated in  Example~\ref{ex:example55}.  \\
Open-source models encountered significant difficulties with \textit{distribution denominator} corrections, frequently becoming confused or trapped in calculation loops. They often attempted to calculate denominators prematurely, as shown in Example~\ref{ex:example56}, revealing poor understanding of the sequential calculation steps required in inheritance law.

\begin{table*}[!h]

\centering
\footnotesize
\renewcommand{\arraystretch}{1.2}
\rowcolors{2}{rowgray}{white}
\begin{tabular}{>{\bfseries}c p{4.7cm} p{9cm}}
\toprule
\# & Question & Answer \\
\midrule

\refstepcounter{examplectr}\label{ex:example55}\theexamplectr 
& \textAR{توفي عن: (أخ شقيق) العدد 3، (ابن أخ شقيق) العدد 3، (بنت) العدد 4، (زوجة) العدد 1، (أخت شقيقة) العدد 5. 
كم عدد أسهم الزوجة من أصل التركة المصحح؟ } 
& \textAR{للزوجة الثمن لوجود الفرع الوارث (البنات)، لقوله تعالى: ﴿فَإِن كَانَ لَكُمْ وَلَدٌ فَلَهُنَّ الثُّمُنُ مِمَّا تَرَكْتُم﴾. أصل المسألة (قبل التصحيح) يكون من مضاعف مقامي نصيب الزوجة (8) ونصيب البنات (3)، وهو $24$. سهم الزوجة من هذا الأصل هو الثمن، أي $\frac{1}{8}$ من $24$ يساوي 3. (ملاحظة: المسألة تحتاج تصحيحاً لوجود باقٍ للعصبة لا ينقسم على رؤوسهم، ولكن هذا التعليل يوضح كيفية الوصول إلى أحد الخيارات المعطاة وهو سهم الزوجة قبل التصحيح).  } \\

\refstepcounter{examplectr}\label{ex:example56}\theexamplectr 
& \textAR{توفي عن: (أخ شقيق) العدد 3، (ابن أخ شقيق) العدد 3، (بنت) العدد 4، (زوجة) العدد 1، (أخت شقيقة) العدد 5. 
كم مجموع عدد أسهم البنات الأربع من أصل التركة المصحح؟ } 
& \textAR{ مجموع عدد أسهم البنات الأربع من أصل التركة المصحح هو $30$ سهمًا. قال الله تعالى في سورة النساء: ﴿وَأُولُوا الْأَرْحَامِ بَعْضُهُمْ أَوْلَى بِبَعْضٍ فِي كِتَابِ اللَّهِ مِنَ الْمُؤْمِنِينَ وَالْمُهَاجِرِينَ إِلَّا أَنْ تَفْعَلُوا إِلَى أَوْلِيَائِكُمْ مَعْرُوفًا كَانَ ذَٰلِكَ فِي الْكِتَابِ مَسْطُورًا﴾ (الآية 7) } \\

\refstepcounter{examplectr}\label{ex:example58}\theexamplectr 
& \textAR{توفي عن زوجة واحدة وأخت شقيقة واحدة، ما هو النصيب النهائي لكل وارث من التركة؟ } 
& \textAR{  الزوجة ترث الربع لعدم وجود الفرع الوارث لقوله تعالى: "فَإِن لَّمْ يَكُن لَّكُمْ وَلَدٌ فَلَهُنَّ الرُّبُعُ مِمَّا تَرَكْتُم". والأخت الشقيقة الواحدة ترث النصف إذا انفردت ولم يكن هناك أصل وارث ذكر ولا فرع وارث، لقوله تعالى: "إِنِ امْرُؤٌ هَلَكَ لَيْسَ لَهُ وَلَدٌ وَلَهُ أُخْتٌ فَلَهَا نِصْفُ مَا تَرَكَ". } \\

\refstepcounter{examplectr}\label{ex:example57}\theexamplectr 
& \textAR{ توفيت عن زوج وأب وأم وبنت، إلى كم تعول هذه التركة؟} 
& \textAR{  في هذه المسألة، المسألة تُعال إلى $12$، وذلك لأن المسألة تُعال إلى $12$ في حال وجود الزوج والأب والأم والبنت، حيث يُضاف الزوج إلى المسألة، فيصبح عدد الرؤوس $5$ (3 + 1 + 1) وتُقسم التركة على 5. قال الله تعالى في سورة النساء: "فَرِيضَةً مِّنَ اللَّهِ إِنَّ اللَّهَ كَانَ عَلِيمًا حَكِيمًا" (النساء: 11) } \\

\refstepcounter{examplectr}\label{ex:example59}\theexamplectr 
& \textAR{توفيت عن زوج وأبناء وبنات هل يجوز للأم تفضيل أحد أولادها في الميراث؟ } 
& \textAR{  لأن قسمة الميراث فريضة مقدرة من الله سبحانه وتعالى بنصوص قطعية في القرآن الكريم، ولا يجوز لأحد تغيير هذه القسمة أو تفضيل بعض الورثة على بعض فيها. قال تعالى في ختام آيات المواريث: (فريضة من الله إن الله كان عليماً حكيماً) [النساء: 11]. } \\

\bottomrule
\end{tabular}
\caption{Examples of Gemini's Complex Errors.}
\label{tab:model_examples3}

\end{table*}

\vspace{0,4em}
\noindent\textbullet\ \textbf{Redistribution (\textit{radd}):} 
Gemini occasionally failed to redistribute leftover shares after the initial allocation, as illustrated in Example~\ref{ex:example58}. In that case, the full sister initially received 50\% of the shares, with 25\% of the estate remaining. This  should go to the heirs other than spouses—in this case, the full sister—bringing her total share to 75\%.  
Open-source models consistently failed in such cases, largely because initial distribution errors made it impossible to identify residual shares requiring redistribution.  

\vspace{0,4em}
\noindent\textbullet\ \textbf{Proportionate Reduction (\textit{ʿawl}):} 
Gemini consistently handled proportionate reduction scenarios accurately, clearly explaining the underlying principle. Conversely, smaller models consistently failed in proportionate reduction calculations, indicating a fundamental misunderstanding of this critical inheritance concept, even when the need for ʿAwl was explicitly stated in the question.  ALLaM  in Example~\ref{ex:example57}, arbitrarily  defined  the ʿAwl denominator without first computing the individual shares  and their common denominator, leading to an invalid final distribution. 

\vspace{0,4em}
\noindent\textbullet\ \textbf{Resolving Exceptional and Disputed Cases:} 
\\
In cases involving juristic disagreement,  all models consistently defaulted to the majority opinion, likely due to a training data bias that favors definitive answers over nuanced scholarly disagreement (\textit{ikhtilāf}). Consequently, they failed to acknowledge valid minority views, as shown in Example~\ref{ex:example59}. The optimal response would be to note Imam Ahmad's less common but valid view: parents may give more to certain heirs during their lifetime—such as those who are very poor or ill—even though this will ultimately impact the total estate. 

\begin{table*}[!h]

\centering
\footnotesize
\renewcommand{\arraystretch}{1.2}
\rowcolors{2}{rowgray}{white}
\begin{tabular}{>{\bfseries}c p{4.5cm} p{8.5cm} >{\bfseries}l}
\toprule
\# & Question & Answer & Model \\
\midrule

\refstepcounter{examplectr}\label{ex:example90}
\theexamplectr & \textAR{توفي عن:  أم، زوجة، أخوين شقيقين، خمس أخوات شقيقات، ثلاثة إخوة من الأب، وأخت من الأب. ما نصيب الأم؟} & \textAR{وفقاً لمعلوماتي، فإن للأم السدس فرضاً لوجود الفرع الوارث، كما قال تعالى: ﴿وَلِأَبَوَيْهِ لِكُلِّ وَاحِدٍ مِنْهُمَا السُّدُسُ مِمَّا تَرَكَ إِنْ كَانَ لَهُ وَلَدٌ﴾ [النساء: 11]. وبالتالي، يكون نصيب الأم في هذه الحالة هو السدس.} & Fanar \\

\refstepcounter{examplectr}\label{ex:example91}
\theexamplectr & \textAR{توفي عن 4 أخ شقيق، 1 أخ من الأم، 1 أم، 4 بنت، 1 زوجة، 2 أخت شقيقة. ما نوع إرث الأخ الشقيق؟} & \textAR{التعليل: يرث الأخ الشقيق بالتعصيب إذا لم يكن هناك وارث ذكر أقرب منه، كما جاء في الحديث النبوي: "الأخ الشقيق يرث ميراث الأخ الشقيق"، وبالتالي يرث الأخ الشقيق بالتعصيب إذا لم يكن هناك وارث ذكر أقرب منه.} & LLaMA \\

\refstepcounter{examplectr}\label{ex:example92}
 \theexamplectr & \textAR{  توفي عن: أخ شقيق (العدد 2)، بنت (العدد 4)، أخت شقيقة (العدد 1). ما هو نصيب البنات الأربع؟}   & \textAR{يرث الأخ الشقيق والأخت الشقيقة والأبناء، فالأخ الشقيق يأخذ سهم الأب (النصف) والأخت الشقيقة تأخذ النصف، وبالتالي يبقى للبنات الثلثان (النصف الباقي) لأن النصاب هو الورثة، والأخ الشقيق والأخت الشقيقة يأخذون نصيبهما، وتبقى البنات على الباقي.  } & LLaMA  \\

\bottomrule
\end{tabular}
\caption{Examples of correct answers accompanied by incorrect justifications.}
\label{tab:model_examples_clean}

\end{table*}

\section{Analysis of Correct Answers}
\label{section4bis}

While standard evaluation focuses on answer accuracy, this is insufficient in religious legal contexts, particularly in inheritance law, where the validity of the underlying reasoning is important. A correct prediction alone may mask significant reasoning flaws if the accompanying explanation is inadequate, irrelevant, or fabricated.
To further investigate the robustness of the models’ reasoning, we manually analyzed the justifications of 40 instances where models produced correct answers. 
Our analysis revealed that even when lower-performing models arrived at the correct answer, their justifications often reflected the same errors discussed in section \ref{eva}.  
In terms of comprehension,  Fanar presumed the presence of offspring, although none were mentioned in the question, as shown in Example~\ref{ex:example90}. Justification errors were also common; models cited irrelevant or fabricated Quranic verses and hadiths, such as LLaMA referencing a non-existent Hadith, see Example~\ref{ex:example91}. Moreover, we observed mistakes in the application of normative rules. In 
Example~\ref{ex:example92}, LLaMA split half the shares between brothers and sisters and gave two-thirds to daughters, while the correct approach is to grant daughters a fixed two-thirds share and allocate the remainder to brothers and sisters.
\\
As shown in Table~\ref{tab:error_categories_summary2}, foundational errors in comprehension and rule application were present in  $27.5\%$ of ALLaM’s justifications, $32.5\%$ of Fanar’s, and $47\%$ of LlaMA’s. In stark contrast, Gemini consistently produced sound justifications, demonstrating accurate calculations, valid legal reasoning, and correct citations. This disparity underscores that performance evaluations must account for reasoning quality, as accuracy alone provides an incomplete and potentially misleading assessment of a model's capabilities in this domain.

 \vspace{-0.1cm}
 
 \begin{table}[!htbp]
\centering
\small
\renewcommand{\arraystretch}{1.3}
\rowcolors{2}{rowgray}{white}
\begin{tabular}{>{\bfseries}lccc}
\toprule
\textbf{Type} & \textbf{ALLaM} & \textbf{Fanar} & \textbf{LLaMA} \\
\midrule
\rowcolor{cyan!15} \textbf{ENR} &22.5 \%   &  22.5\% & 44.5\% \\
\rowcolor{cyan!15} \textbf{CE} &  5\% & 10\% & 2.5\%   \\
\rowcolor{cyan!15} \textbf{BCE} & - & - & - \\
\rowcolor{cyan!25} \textbf{FD Total} & 27.5\% & 32.5\% &  47\%\\
\midrule
\rowcolor{red!15} \textbf{ECA} & - & - &  -\\
\rowcolor{red!25} \textbf{CPLX Total} & - & - & - \\
\bottomrule
\end{tabular}
\caption{Distribution of error types in model justifications for correct answers. }
\label{tab:error_categories_summary2}
\vspace{-0.5em}
\end{table}


\section{Conclusion}
 \label{sect05}
 
This paper addresses estate distribution according to Islamic inheritance law using seven distinct LLMs. Due to the task’s complexity, models with reasoning capabilities, such as Gemini 2.5 and o3, demonstrated high performance,  achieving accuracy rates of 90.6\% and 93.4\%, respectively.  Models without reasoning capability, such as GPT-4.5—which is considered one of the most powerful commercial OpenAI models—achieved moderate results (74\%). Conversely, models like Jais, Mistral, and LLaMA, despite strong performance on several Arabic language benchmarks, showed significantly lower accuracy, scoring below 50\%, reflecting their limitations in legal reasoning.
Our evaluations highlighted a clear gap between models with reasoning abilities and those without. This gap was particularly evident among ALLaM, Fanar, LLaMA, and Mistral, which consistently struggled with identifying complex familial relationships, evaluating diverse inheritance scenarios, and correctly executing corrective calculations such as redistribution (\textit{Radd}) and proportionate reduction (\textit{ʿAwl}). Future research should focus on developing agentic AI systems that can reason step by step, follow legal rules precisely, and adapt to complex inheritance cases. This includes creating high-quality datasets designed to support legal reasoning, developed with the guidance of experts in Islamic law. 

\bibliography{custom}

\end{document}